\documentclass{article}
\usepackage{spconf}
\usepackage{amsmath}
\usepackage{graphicx}

\title{A Robust Contrastive Alignment Method for Multi-domain Text Classification}
    \name{Xuefeng Li$^{\dagger, *}$\thanks{\ \ ${^{*}}$The first two authors contribute equally.} Hao Lei$^{\dagger, *}$ 
    Liwen Wang${^{\dagger}}$
    Guanting Dong${^{\dagger}}$
    Jinzheng Zhao$^{\ddagger}$
    Jiachi Liu${^{\dagger}}$
    \textit{Weiran Xu}${^{\dagger,\star}}$\thanks{ ${^{\star}}$Weiran Xu is the corresponding author.}
    \textit{Chunyun Zhang}${^{\S}}$}
\address{ ${^{\dagger}}$Beijing University of Posts and Telecommunications, Beijing, China \\
          ${^{\ddagger}}$School of Computer Science and Electronic Engineering, University of Surrey, UK\\
          ${^{\S}}$Shandong University of Finance and Economics, Jinan, China}
\begin{document}
\maketitle

\begin{abstract}
Multi-domain text classification can automatically classify texts in various scenarios. Due to the diversity of human languages, texts with the same label in different domains may differ greatly, which brings challenges to the multi-domain text classification. Current advanced methods use the private-shared paradigm, capturing domain-shared features by a shared encoder, and training a private encoder for each domain to extract domain-specific features. However, in realistic scenarios, these methods suffer from inefficiency as new domains are constantly emerging. In this paper, we propose a robust contrastive alignment method to align text classification features of various domains in the same feature space by supervised contrastive learning. By this means, we only need two universal feature extractors to achieve multi-domain text classification. Extensive experimental results show that our method performs on par with or sometimes better than the state-of-the-art method, which uses the complex multi-classifier in a private-shared framework.
\end{abstract}

\vspace{-0.1cm}
\section{Introduction}
\label{sec:intro}

Text classification is widely applied in many scenarios, such as topic classification \cite{wang2012baselines}, intention recognition \cite{devlin2018bert} and emotion detection \cite{wang-zong-2021-distributed}. Existing data-driven emotion detection methods have achieved high accuracy in many scenarios. However, the performance of these methods depends on numerous annotated data in the domain. When applying to a new domain,
the model has difficulty making predictions precisely for the new data distribution. Therefore, it is of great significance to design a multi-domain text classification model to improve the robustness of the model in different domains.

To address the problem of multi-domain text classification, a straightforward way is to directly mix all the data of different domains and train a single model.
\begin{figure}[t]
  \centering
  \centerline{\includegraphics[width=9.5cm]{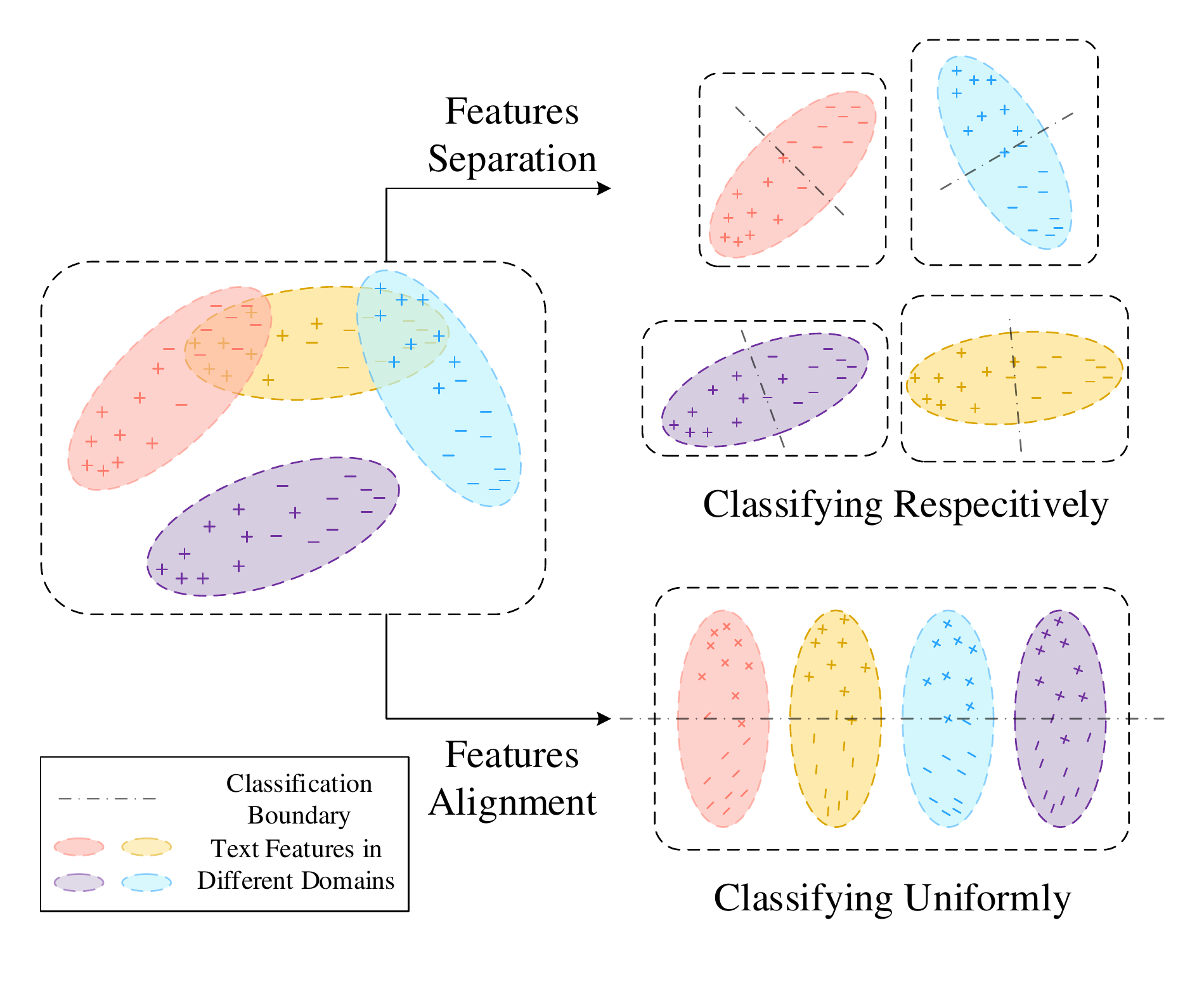}}
  \vspace{-0.8cm}
%
\caption{The left represents the original multi-domain text feature space, where exists a domain feature mismatch problem. The top right represents the shared-private paradigm, which performs multi-domain text classification tasks independently through multiple classifiers. And the bottom represents our method, where text features of different domains are aligned in the same feature space, and a single classifier is used to complete multi-domain classification.}
\vspace{-0.4cm}
\label{fig:res}
\end{figure}

This enables the model to learn the classification information from multiple domains. However, there is a risk of feature mismatch between the distribution of text categorization information in different domains. 
It is difficult for a single classifier to distinguish the classification features of all domains in the same feature space. Therefore, a private-shared paradigm \cite{7373350}\cite{liu2017adversarial}\cite{chen2018multinomial}\cite{8852388}\cite{wu2021mixup}\cite{wu2020dual}\cite{wu2021conditional} is applied to this task. These methods learn a private feature extractor specifically for each domain, while learning a shared feature extractor for all domains. The private feature extractor is used for texts of each domain separately, and the shared extractor is used to extract general information in all domains to improve the classification accuracy. Although many improvements have been made, it is not realistic to learn a model for each of the new domains that are constantly emerging in realistic scenarios.

Inspired by the above challenge, in this paper, we propose a \textbf{R}obust \textbf{C}ontrastive \textbf{A}lignment method (\textbf{RCA}) for the multi-domain text classification task, which only contains one domain information extractor and one classification feature extractor. First, since all domains use the same classifier, we need to solve the problem of multi-domain classification alignment. We use supervised contrastive learning of domain information to train the domain information extractor to encourage the separation of multi-domain text features and prevent the confusion and overlap of different domain classification features. Then through supervised contrastive learning based on classification information, the classifier information extractor is trained to force the text features of each domain to align with each other according to the classification information. As shown in figure \ref{fig:res}, after the classification features of different fields are aligned in the same feature space, the multi-domain text classification can be accomplished by a single classifier. In addition, due to the great diversity of multi-domain text distribution, it is difficult to learn the parameters of a single model at the same time even after alignment. Therefore, we introduce anti-attack training to improve the robustness and generalization ability of the model. Our main contributions are three-fold: (1) We propose a Contrastive Alignment Method which only needs a universal model for multi-domain text classification; (2) We introduce an adversarial attack training strategy to improve model robustness. (3) Extensive experimental results show that our proposed RCA method performs on par with or sometimes better than the state-of-the-art method, which uses the complex multi-classifier in a shared-private framework.


\section{PROPOSED METHOD}
\label{sec:format}
Before describing our method, we first give the definition of the task. The MDTC task is a classification task oriented to real-world scenarios where texts come from different domains. Suppose there are $M$ domains $\{D_i\}^{M}_{i=1}$ and each domain has $L_i$ labeled data $\{(x_j,y_j)\}_{j=1}^{L_i}$. We expect to train a classifier to assign the correct label $y_j$ to an instance $x_j$. We use the average classification accuracy rate in all domains to evaluate our method.  It is worth mentioning that the previous methods train a unique classifier for each domain, while we  use a universal classifier to test on all domains.

\subsection{Model Architecture}
\label{ssec:subhead}
\begin{figure}[t]
  \centering
  \centerline{\includegraphics[width=10cm]{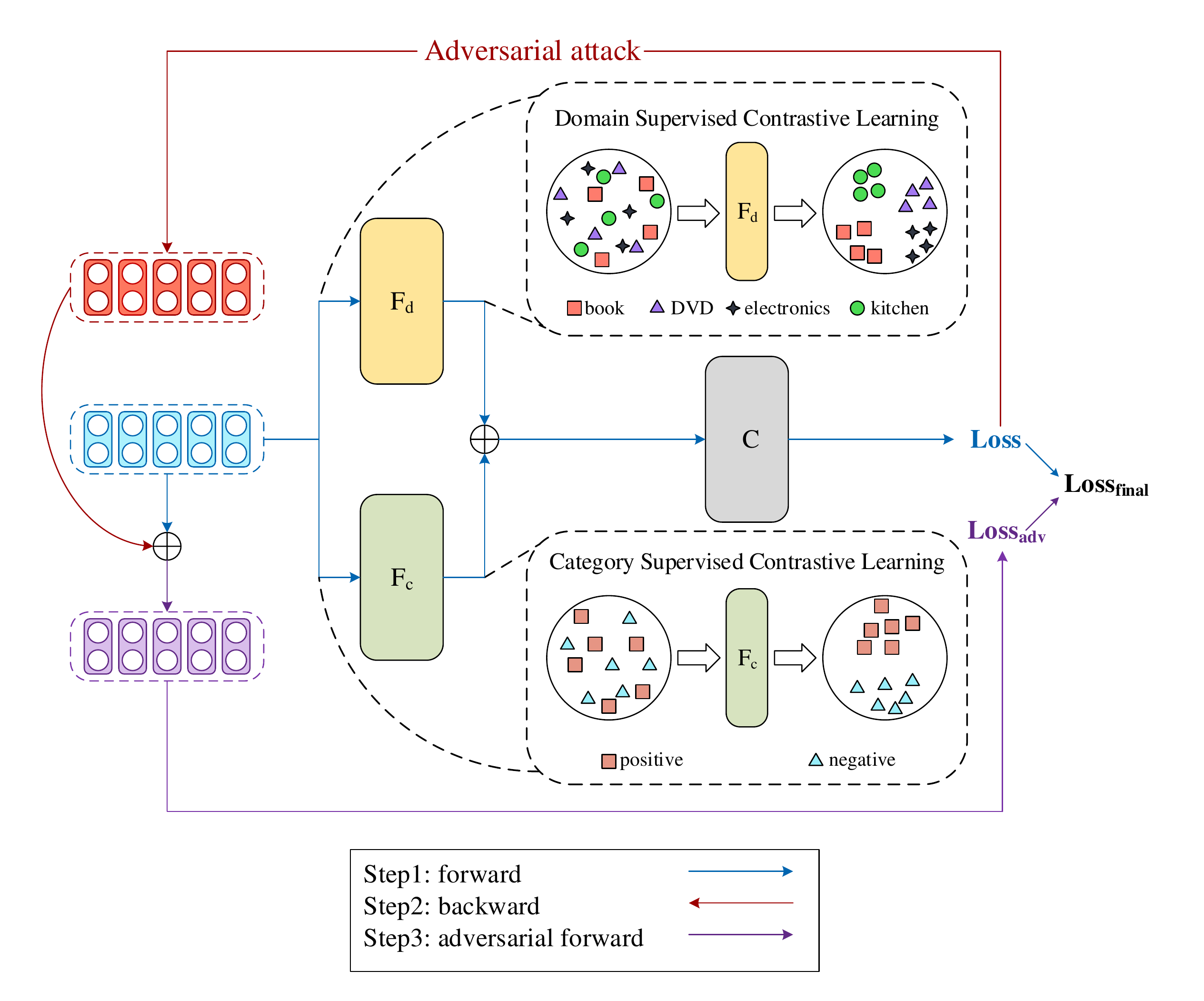}}
  \vspace{-0.3cm}
%

\caption{The overall architecture of our method, including three steps: forward, backward and adversarial forward. First, the domain feature extractor $F_d$ and the category  feature extractor $F_c$ encode the instances, and the joint features are sent to the classifier $C$ to conduct text classification. Then the original loss is calculated and backward. Finally, the noise disturbs the input embedding to get the new input, which performs the adversarial forward to calculate the final loss. }
\label{fig:re}
\end{figure}

As shown in Figure \ref{fig:re}, our model includes three parts: a domain feature extractor $F_d$, a category feature extractor $F_c$ and a sentiment classifier $C$.
For each sentence $x$, it first passes $F_d$  to get the domain feature $f_d=F_d(x)$, and passes $F_c$ to get the category feature $f_c=F_c(x)$. We concatenate the two features and get the final representation of the sentence $h=concat(f_d,f_c)$. The classifier $C$ and a softmax layer then map $h$ to the predictor label $\tilde y$ . 

We use domain supervised contrastive learning to distinguish confusing domain features, and use category supervised contrastive learning to align the category features of samples. In addition, for some representations near the classification boundary, the classifier is often difficult to classify accurately. To solve the problem, we introduce adversarial attack during training to force the model to learn more robust representations.

\subsection{Domain Supervised Contrastive Learning}
\label{ssec:subhead}
To separate the text features in different domains, we apply the supervised contrastive learning \cite{DBLP:journals/corr/abs-2004-11362} \cite{wang-etal-2021-bridge}\ to train the domain feature extractor. We consider the samples from the same domain as positive samples and samples from other domains as the negative samples to keep the text feature of the same domain together and away from other features.

For $N$ instances$\{x_i^j\}_{i=1}^N$ from different domains, $x_i^j$ is the $i$-th instance from domain $D_j$. We take instances $x_i$ from the same domain as positive examples:
\begin{equation}
    x_{i,d}^+=\{x_k^j\}_{k=1}^m, k\neq i
\end{equation}
while the other instances $x_{i,d}^-$ from different domains as the negative examples, $m$ represents the number of positive samples in a batch. Then we train $F_d$ by minimizing the InfoNCE loss as follows \cite{aitchison2021infonce}\cite{tsai2021conditional}:
\begin{equation}
    l_d = -\frac{1}{N}\sum_{i=1}^Nlog(\frac{exp(S(F_d(x_i),F_d(x_{i,d}^+))/\tau_1)}{\sum_{j=1}^{N-1}{exp(S(F_d(x_i),F_d(x_j))/\tau_1)}})
\end{equation}
where $S(a,b)$ represents the similarity function that measures the distance between $a$ and $b$. We use cosine similarity $S(\cdot, \cdot)$ in our experiments. $\tau_1$ is the temperature coefficient.

\subsection{Category Supervised Contrastive Learning}
\label{ssec:subhead}
As we have to classify all text features from different domains in a same feature space, we need to align text features from different domains in the same feature space. Similarly, we use category contrastive learning to help $F_c$ learn better category representations. We use instances of the same category as $x_{i}$  as positive samples $x_{i,c}^+$, and the other instances as negative samples. The loss of $F_c$ is similar to $l_d$:
\begin{equation}
    l_c = -\frac{1}{N}\sum_{i=1}^Nlog(\frac{exp(S(F_c(x_i),F_c(x_{i,c}^+))/\tau_2)}{\sum_{j=1}^{N-1}{exp(S(F_c(x_i),F_c(x_j))/\tau_2)}})
\end{equation}
where $\tau_2$ is the temperature coefficient.

Domain contrastive learning and category contrastive learning make the current instance distinguish from samples of different domains and different categories in the feature space. The representation with both domain and category information will make the classification more versatile and accurate.

\subsection{Adversarial Learning}
\label{ssec:subhead}
In order to enhance the robustness of the model, we introduce adversarial training \cite{yan2020adversarial}\cite{wang-etal-2021-dynamically} to reduce the influence of domain noise. First, $x_i$ is encoded to obtain the representation $h_i$, and the negative log-likelihood (NLL) loss is calculated as follows:
\begin{equation}
    l_C = -\frac{1}{N}\sum_{i=1}^Nlog(softmax(C(h_i)))
\end{equation}
We then backward the above classification loss to get the noise $\widetilde v_{noise}$ added to the input embedding:
\begin{equation}
    \widetilde v_{noise} = \epsilon \frac{g}{\left \|g\right \|},g = \nabla_el_C
\end{equation}
where $\epsilon$ is the normalized noise coefficient. We perform the second forward to get a new loss $l_C^{\prime}$, and the new classification loss $l_{adv}= (1-\lambda)l_C + \lambda l_C^{\prime}$. Finally, we optimize the entire model by minimizing $l_{total}$:
\begin{equation}
    l_{total} = l_{adv}+\alpha (l_d+l_c)
\end{equation}
where $\lambda$ and $\alpha$ are the hyperparameters for balancing different losses.

\begin{table}[t!]
\centering
\resizebox{85mm}{30mm}{
 \begin{tabular}{|c c c c c c c c|} 
 \hline
 Domain & CNN-single & ASP-MTL & MAN-L2 & MAN-NLL & MT-BL & MRAN & RCA \\ [0.5ex] 
 \hline
  books & 81.75 & 84.0 & 87.6 & 86.8 & \textbf{89.0} & \underline{87.0} & 86.75 \\ 
 electronics & 84.5 & 86.8 & 87.4 & 88.8 & \textbf{90.2} & \underline{89.0} & 88.25 \\ 
 dvd & 83.25 & 85.5 & 88.1 & 88.6 & 88.0 & \underline{89.0} & \textbf{89.75} \\ 
 kitchen & 84.75 & 86.2 & 89.8 & 89.9 & \underline{90.5} & \textbf{93.0} & 90.0 \\ 
 apparel & 82.5 & 87.0 & \underline{87.6} & \underline{87.6} & 87.2 & \textbf{91.5} & 87.5 \\ 
 camera & 85.25 & 89.2 & 91.4 & 90.7 & 89.5 & \textbf{93.0} & \underline{91.75} \\ 
 health & 85.25 & 88.2 & 89.8 & 89.4 & \underline{92.5} & 90.0 & \textbf{90.75} \\ 
 music & 82.5 & 82.5 & 85.9 & 85.5 & 86.0 & \underline{86.5} & \textbf{87.0} \\ 
 toys & 86.5 & 88.0 & 90.0 & \textbf{90.4} & 92.0 & 86.0 & \underline{90.25} \\ 
 video & 86.5 & 84.5 & 89.5 & \underline{89.6} & 88.0 & 88.5 & \textbf{90.75} \\ 
 baby & 80.75 & 88.2 & 90.0 & \underline{90.2} & 88.7 & 90.0 & \textbf{91.0} \\ 
 magazine & 85.5 & 92.2 & 92.5 & 92.9 & 92.5 & \textbf{93.5} & \underline{92.75} \\ 
 software & 84.5 & 87.2 & 90.4 & \underline{90.9} & \textbf{91.7} & 89.5 & 88.5 \\ 
 sports & 86.25 & 85.7 & 89.0 & 89.0 & 89.5 & \textbf{90.5} & \underline{90.25} \\ 
 IMDb & 83.25 & 85.5 & 86.6 & 87.0 & 88.0 & \underline{89.0} & \textbf{90.5} \\ 
 MR & 72.75 & 76.7 & 76.1 & 76.7 & 75.7 & \textbf{78.5} & \textbf{78.5} \\ 
 \hline
 AVG & 84.75 & 86.1 & 88.2 & 88.4 & 88.6 & \textbf{89.0} & \textbf{89.0} \\ [1ex] 
 \hline
 \end{tabular}}
 \caption{MDTC classification accuracies on the FDU-MTL dataset. Bold font denotes the best classification accuracy results and underline denotes the second-best results.}
 \label{table:1}
\end{table}


\begin{table}[t!]
\centering
\resizebox{85mm}{10mm}{
 \begin{tabular}{|c c c c c c c c c|} 
 \hline
 Domain & CNN-single & CMSC-LS & CMSC-SVM & CMSC-Log & MAN-L2 & MAN-L2 & MRAN & RCA \\ [0.5ex] 
 \hline
 Books & 82.95 & 82.10 & 82.26 & 81.81 & 82.46 & 82.98 & \underline{84.60} & \textbf{85.50} \\ 
 Electronics & 83.10 & 82.40 & 83.48 & 83.73 & 83.98 & 84.03 & \textbf{85.60} & \underline{84.50} \\ 
 dvd & 86.25 & 86.12 & 86.76 & 86.67 & 87.22 & 87.06 & \textbf{89.10} & \underline{88.50} \\ 
 kitchen & 83.10 & 82.40 & 83.48 & 83.73 & 83.98 & 84.03 & \textbf{85.60} & \underline{84.50} \\ 
 \hline
 AVG & 84.96 & 84.55 & 85.18 & 85.11 & 85.55 & 85.66 & \textbf{87.64} & \underline{86.88} \\ 

 \hline
 \end{tabular}}
 \caption{MDTC classification accuracies on the Amazon review dataset. Bold font denotes the best classification results and underline denotes the second-best results.}
 \label{table:2}
\end{table}

\section{EXPERIMENTS}
\label{sec:typestyle}

\subsection{Dataset}
\label{ssec:dataset}

Two benchmark datasets are used in our experiment: Amazon review dataset\cite{blitzer2007biographies} and FDU-MTL dataset\cite{liu2017adversarial}.
Amazon review dataset includes 4 domains, namely books, DVDs, electronics and kitchen. Each domain consists of 1000 positive and 1000 negative, a total of 2000 samples. Since the dataset is pre-processed into a bag of features (unigrams and bigrams) and loses all word order information, we take the 5,000 most frequent features and represent each comment as a 5,000-dimensional vector.
The FDU-MTL contains 16 domains, two of which are movie review datasets, namely IMDB and MR. The other 14 domains are
Amazon product reviews, including books, electronics, DVDs, kitchen, apparel, camera, health, music, 
toys, video, baby, magazine, software, and sport. The dataset is raw text data and is tokenized by
Stanford tokenizer.

\subsection{Comparison Methods}
\label{ssec:methodcom}

In order to better evaluate the performance of our method, we list several baselines for comparison. We propose a simple baseline CNN-single which uses CNN as the feature extractor and MLP as the classifier. It trains and tests on data from all domains.
The collaborative multi-domain sentiment classification (CMSC) combines a multi-domain classifier
and a separate classifier for each domain to make predictions\cite{7373350}.
The CMSC models can be trained on three different loss functions: the least square loss (CMSC-LS), the hinge loss (CMSC-SVM) and the log loss (CMSC-Log).
The adversarial multi-task learning for text classification (ASP-MTL)\cite{liu2017adversarial}. It adopts a share-private 
scheme and introduces adversarial learning and orthogonality constraints to ensure that the shared feature space contains only shared information.
Multinomial Adversarial Networks (MANs) exploits two forms of loss functions to train the multinomial domain discriminator: the least square loss (MAN-L2) and the negative log-likelihood loss (MAN-NLL)\cite{chen2018multinomial}.
Multi-task Learning with Bidirectional Language Models (MT-BL) adds a language model to
domain-specific feature extraction, which can better extract domain-specific information\cite{8852388}. A uniform label distribution is used on the feature shared extractor.
Mixup Regularized Adversarial Networks (MRAN) adds a mix-up to the data input part, combines labeled data and unlabeled data, and improves task performance through data enhancement\cite{wu2021mixup}.
\vspace{-0.5cm}
\subsection{Implementation Details}
\label{ssec:details}
In the experiment on the Amazon review dataset, the dataset only has feature embedding and 
lacks word order, so we use MLP as the feature extractor. Same as the previous settings, each sentence 
takes a 5000-dimensional vector as the final representation and input to the feature extraction. 
The dimension of the feature embedding output by the MLP is 128-dimension, so the input dimension 
of the classifier $C$ is 128+128 dimensions.
In the experiment on the FDU-MTL dataset, we use CNN as the feature extractor. The size of the
convolution kernel used is (3, 4, 5), and the number is 200. The 100-dimensional word embedding 
of \emph{word2vec} is used as the word vector, the dimensional size of the CNN output is still
128-dimensional, and the other settings are consistent with the Amazon review dataset.

We set the hyperparameters used in the experiment as follows: $\tau_{1}$ = $\tau_{2}$ = 0.1, 
$\epsilon$ = 0.3, $\lambda$ = 0.3, $\alpha$ = 0.01. In the experiment, we set 
the batch size as 32. We use Adam as the optimizer, the learning rate is set to 0.0001, 
and the dropout rate is set to 0.4.

\subsection{Results}
\label{ssec:details}
The main results are shown in Table \ref{table:1} and Table \ref{table:2}. Our universal model obtains SOTA results on FDU-MTL dataset, and also performs better than most multi-classifier models except MRAN on Amazon dataset. This is mainly because MRAN used numerous unlabeled data, while we only use annotated data. Compared to the single classifier baseline CNN-single, our model improves by 2.92\% on Amazon dataset and 4.25\% on FDU-MTL dataset respectively, which fully demonstrates the effectiveness of our framework. The domain and categorical features cannot be well represented by a single encoder, so the effect on multi-domain tasks is not good. Our approach uses contrastive learning to align domain information and categorical information to learn more effective representations. In addition, we introduce adversarial training to make the instances at the classification boundary closer to the class center, which enhances the robustness and generalization ability of the model.
\begin{table}[t!]
\centering
\resizebox{70mm}{13mm}{
 \begin{tabular}{|c | c | c | c | c | c|} 
 \hline
 Method & Amazon & FDU-MTL & AVG\\ [0.5ex] 
 \hline
RCA(full) & 86.88 & 89.01 & 87.95\\
RCA w/o DSCL  & 85.62 & 88.34 & 86.98\\ 
RCA w/o CSCL  & 85.50 & 88.56 & 87.03\\
RCA w/o AL & 85.56 & 88.35 & 86.96\\ [1ex] 
 \hline
 \end{tabular}}
 \caption{Ablation study analysis.}
 \vspace{-0.4cm}
 \label{table:3}
\end{table}

\subsection{Ablation Study}
\label{ssec:details}
To verify the validity of each part of our framework, we performed ablation experiments on Amazon and FDU-MTL dataset. As shown in Table \ref{table:3}, removing all three parts will result in varying degrees of decline in model performance. This is consistent with our conjecture, since domain supervised contrastive learning (DSCL) and category supervised contrastive learning (CSCL) play a role in feature alignment together, missing one part will make the representation more chaotic, and thus make the classification effect worse. At the same time, the experiment also illustrates the effect of adversarial learning (AL), which makes the classification boundary clearer and enhances the classification ability of the model for easily confused samples.


\section{CONCLUSION}
In this paper, we propose a robust contrastive alignment method for multi-domain text classification. We use domain supervised contrastive learning to prevent domain information confusion and category supervised contrastive learning to align categorical information of different texts. At the same time, we introduce adversarial attack training to enhance the model's robustness and generalization ability. Compared with the previous method, our method is simpler and more effective. Experiments on two MDTC benchmarks prove the effectiveness of our proposed methods.

\section{Acknowledgements}
This work was partially supported by National Key R\&D Program of China No. 2019YFF0303300 and Subject II  No. 2019YFF0303302, DOCOMO Beijing Communications Laboratories Co., Ltd, MoE-CMCC "Artifical Intelligence" Project No. MCM20190701.

\label{sec:typestyle}


\vspace{1cm}
\bibliographystyle{IEEEbib}
\bibliography{strings,refs}

\end{document}